\documentclass{INTERSPEECH2023}

\usepackage{hyperref}
\usepackage{multirow}
\usepackage{xcolor}
\usepackage{subcaption}

\interspeechcameraready

\title{Handling the Alignment for Wake Word Detection: \\A Comparison Between Alignment-Based, Alignment-Free and Hybrid Approaches}
\name{Vinicius Ribeiro$^{1,2}$, Yiteng Huang$^1$, Yuan Shangguan$^1$, Zhaojun Yang$^1$, Li Wan$^1$, Ming Sun$^1$}
\address{
  $^1$Meta AI \\
  $^2$Université de Lorraine, CNRS, Inria, LORIA, F-54000, Nancy, France}
\email{\{ribeirovinicius, yah, yuansg, zhaojuny, wwanli, sunming425\}@meta.com}

\begin{document}

\newcommand{\ie}{i.e.}
\newcommand{\eg}{e.g.}
\newcommand{\quotes}[1]{``#1"}

\maketitle

\makeatletter
\def\blfootnote{\gdef\@thefnmark{}\@footnotetext}
\makeatother
 
\begin{abstract}

Wake word detection exists in most intelligent homes and portable devices. It offers these devices the ability to \quotes{wake up} when summoned at a low cost of power and computing.
This paper focuses on understanding alignment's role in developing a wake-word system that answers a generic phrase. We discuss three approaches. The first is alignment-based, where the model is trained with frame-wise cross-entropy. The second is alignment-free, where the model is trained with CTC. The third, proposed by us, is a hybrid solution in which the model is trained with a small set of aligned data and then tuned with a sizeable unaligned dataset. We compare the three approaches and evaluate the impact of the different aligned-to-unaligned ratios for hybrid training. 
Our results show that the alignment-free system performs better than the alignment-based for the target operating point, and with a small fraction of the data ($20\%$), we can train a model that complies with our initial constraints.


\end{abstract}


\blfootnote{Work was done when Vinicius Ribeiro was an intern at Meta AI. Correspondence to Yiteng Huang: \href{mailto:yah@meta.com}{yah@meta.com}}

\section{Introduction}\label{sec:intro}

Wake word detection, also known in the literature as keyword spotting, refers to identifying if target phrases appear in an audio sequence. 
With the advancement of virtual assistants like Google Assistant, Amazon Alexa, Apple Siri, wake word engines are present in most of the edge devices available in the market, be they phones, tablets, watches, or glasses. 
Wake words work as gateways to these devices. Due to energy consumption constraints, these devices operate most of the time in a low energy consumption state. They are thus not expected to recognize any commands until they register a call to the wake word. 
Once activated, they shift to a high energy consumption state with more powerful computation to completely recognize the user's instructions. 
Wake word detection accuracy is thus essential for a smooth user experience.
To be specific, a system with a high False Accept Rate will trigger too frequently, being annoying to the user and raising concerns about privacy issues~\cite{combs2022you, schonherr2022exploring}. Conversely, a system with a high False Reject Rate will incapacitate the person from using the product. Most often, these two aspects represent a trade-off, \ie, changing the activation threshold towards one metric will inevitably damage the other.

Initially, wake word detection models were developed using Hidden Markov Models, which model both the voice activity and the background noise~\cite{rose1990hidden, wilpon1990automatic}, and later they shifted to deep neural networks~\cite{chen2014small, yang2022lico} as most of the literature in complex data processing did. Wake word detection models are traditionally built on top of two main techniques.
On the one hand, there are alignment-based approaches~\cite{panchapagesan16_interspeech,lopez2021deep}, which assume that the exact alignment between the phoneme targets of the utterances and the corresponding audio are available during model training. Such alignment simplifies the task since the models are trainable with traditional cross-entropy (CE) loss. The alignments, however, obtained using forced alignment algorithms, not only are computationally expensive, but also might introduce annotation errors and be unavailable for low-resource languages~\cite{mathad2021impact}. 
On the other hand, there exist alignment-free techniques~\cite{coucke2019efficient, hou2019region}, with which phonetic alignments of audio transcriptions are not needed during model training. Alignment-free approaches are common in Automatic Speech Recognition (ASR)~\cite{wang2019overview, miao2020online}, commonly used with the Connectionist Temporal Classification (CTC) loss~\cite{graves2012connectionist}. CTC uses dynamic programming to search for the most likely alignment between all possibilities in an efficient manner. The most significant advantage of using alignment-free detection is that it enables the usage of much larger dataset. Additionally, it allows federated learning with edge devices~\cite{wang2019adaptive} since the users' data do not have to leave their devices to be pre-processed. Alternatively, we might have at our disposal some aligned data that we would like to benefit from. The idea proposed by us is that this small dataset could provide a good starting point for the acoustic model of the alignment-free model. We refer to this approach as hybrid alignment. We initially hypothesize that the model trained with a large aligned set would represent a performance upper bound; likewise, the model trained with the small unaligned set would represent a performance lower bound. However, when training the model with a small aligned set using cross-entropy loss, and then continuing the training with a large unaligned set using CTC loss, the model can find an intermediate spot where the performance is improved in comparison to an alignment-based system trained with the smaller share of the corpus and to an alignment-free system trained with the larger share of the corpus.

To our knowledge, few previous works in wake word detection have compared alignment-based and alignment-free approaches under the same setup. Also, this work is the first to propose a hybrid alignment method to benefit from aligned and unaligned data in the same system. Our main contribution is to fill the gap by comparing the three alternatives described under the same conditions and with the same dataset for training and evaluation.
Our second contribution, is to present results that contradicts our initial hypothesis -- that the alignment-based training approach does not represent a performance upper bound, and the CTC-alignment based approach does not represent a lower bound. The best approach should be decided based on the operating point chosen for the target use case.

In this work, we trained a wake word detection model on $274\,194$ utterances, totaling $523$ hours of speech. We explore the amount of utterances with wake words, referred to as \quotes{positive} data, needed to achieve reasonable results with alignment-based model training. Similarly, we identify the amount of positive data for the alignment-free approach. We show that the alignment-based training performs better for a high FAh ($> 0.5$~FAh), while the alignment-free performs better for low levels of FAh. We than combine the two approaches and observe that with a $50\%/50\%$ aligned vis-à-vis unaligned data ratio, the model retains the best of the two approaches.

\section{Data Preparation}\label{sec:dataprep}

The dataset used in this work contains positive samples with 4\,274 unique speakers (2\,945 female and 1\,329 male) collected under the users' agreement via dogfooding and paid recording sessions.
Speakers have unequal contribution to the dataset, meaning that a few speakers supplied hundreds of utterances while many speakers contribute with only a few of them. This imbalance could bias the model towards the most significant contributors, preventing it from generalizing. To handle this imbalance, we hold out speakers with a low contribution (less than 50 utterances) for the final evaluation. Additionally, we limit the individual contribution in training (100 utterances/speaker) and the evaluation set (10 utterances/speaker) such that no speaker has a disproportional amount of samples in the dataset. Finally, we discard positive utterances longer than 20 seconds. The final train and evaluation datasets are organized in a speaker-independent way.

Additionally, we use irrelevant speech that are very unlikely to contain the wake word as negative data. The same negative set was used in all of the experiments, which differed only in the positive data. 
\autoref{tab:dataset} summarizes the number of speakers, utterances, and duration in hours available for train and evaluation.

The phonetic annotations are first obtained by running a non-streaming ASR model to obtain the audios' transcription; then, we run forced alignment on all samples. We augment the positive train set by applying speed distortion and adding background noise. For each original audio sample, five new augmented copies are created. To evaluate how much data is necessary for training each approach, we split the set into several subsets referred to as A and B. We denote set \texttt{A[X]} the dataset containing \texttt{X\%} of the data and \texttt{B[Z]} the dataset containing the remaining samples, \ie, \texttt{A[X]} and \texttt{B[Z]} are complementary to each other ($Z = 100 - X$). In addition, we denote \texttt{CE-A[X]} and \texttt{CE-B[Z]} the alignment-based models trained with the datasets \texttt{A[X]} and \texttt{B[Z]}, respectively. Similarly, we denote \texttt{CTC-A[X]} and \texttt{CTC-B[Z]} the alignment-free models trained with the datasets \texttt{A[X]} and \texttt{B[Z]}, respectively. Finally, we denote \texttt{CE-A[X]-CTC-B[Z]} the alignment-hybrid models trained with \texttt{A[X]} using CE followed by \texttt{B[Z]} using CTC. It is important to highlight that the set \texttt{A[$X_i$]} contains the set \texttt{A[$X_j$]} for all $X_i > X_j$. Likewise, the set \texttt{B[$Z_i$]} contains the set \texttt{B[$Z_j$]} for all $Z_i > Z_j$.

\begin{table}
    \centering
    \caption{Number of speakers, utterances and hours of speech for each dataset.}
    \begin{tabular}{ccccc}
    \toprule
    \multicolumn{2}{c}{Dataset} & Speakers & Utterances & Duration (h) \\
    \midrule
    \multirow{2}{*}{Train} & Positive & 2\,369 & 234\,753 & 216.8 \\
                           & Negative & --     &  39\,441 & 306.0 \\
    \multirow{2}{*}{Eval}  & Positive & 1\,905 &  10\,116 &  10.6 \\
                           & Negative & --     &  16\,635 & 194.0 \\
    \bottomrule
    \end{tabular}
    \label{tab:dataset}
\end{table}

\section{Methods}\label{sec:methods}

Our network is based on the neural network topology called SVDF (single value decomposition filter), first introduced by \cite{nakkiran2015compressing} and discussed in detail in \cite{alvarez2019end}. The network is composed of an encoder that emits per-frame class probabilities while the decoder predicts the occurrence of the wake word. We utilize only the encoding path differently from \cite{alvarez2019end}. Our model takes 80-dimensional log-Mel filter-bank energies computed over a 25~ms window every 10~ms as input and is trained to recognize the nine phonemes of the wake word plus three extra tokens corresponding to silence, unknown and blank (used for CTC). We replaced the decoder network by a decoding rule-based algorithm. We run a sliding window where we expect to observe the wake word through the audio file. The emission probabilities inside the decoding window are smoothed, and the log probabilities are computed. Finally, the best decoding path for a given wake word candidate is selected using the Max Pooling Viterbi algorithm, \ie, instead of summing the log probabilities, the maximum log probability in the sequence of the same token predictions is computed.

The models are trained for 180 epochs. In the hybrid-alignment case, the training is divided such that during the first 90 epochs, the model runs with cross-entropy loss and the following 90 epochs with CTC loss. The models are trained with Adam optimizer~\cite{kingma2014adam} using a weight decay of $1\mathrm{e}{-2}$ and a learning rate of $5\mathrm{e}{-3}$ during the first $60$ epochs and then decreased by a factor of $0.96$ per epoch. For hybrid alignment, the CTC training phase uses a fixed learning rate of $5\mathrm{e}{-4}$. The code was implemented using \texttt{PyTorch}~\cite{NEURIPS2019_9015}.

We evaluate the models in terms of Detection Error Trade-off (DET) curves~\cite{martin1997det}, in which the x-axis represents the number of false alarms per hour (FAh), and the y-axis represents the chance of false rejects per utterance (FRR). User experience research indicates that the users are usually satisfied with an FRR around $5\%$ at the $0.1$ FAh level, even though an FRR of $10\%$ is acceptable for many use cases. Therefore we evaluate the models' performances at this level. In addition, we measure the latency introduced by CTC relative to the alignment-based method. It is important to stress that we are not measuring the latency between the prediction and the actual occurrence of the wake word but the difference in the triggering points of alignment-based and alignment-free engines. To perform such a calculation, for each utterance, we get the peak value of each system and compute the point where the decoder score is greater than $40\%$ of the peak value. We force this threshold to be greater or equal to $0.20$ to guarantee that we are not computing cases where the system did not fire. Then we calculate the difference between the triggering point for cross entropy and CTC -- positive measures mean that the alignment-based system triggered first, and negative measures represent the opposite.

\section{Results}\label{sec:results}

\begin{figure*}[t!]
    \centering
    \includegraphics[scale=1]{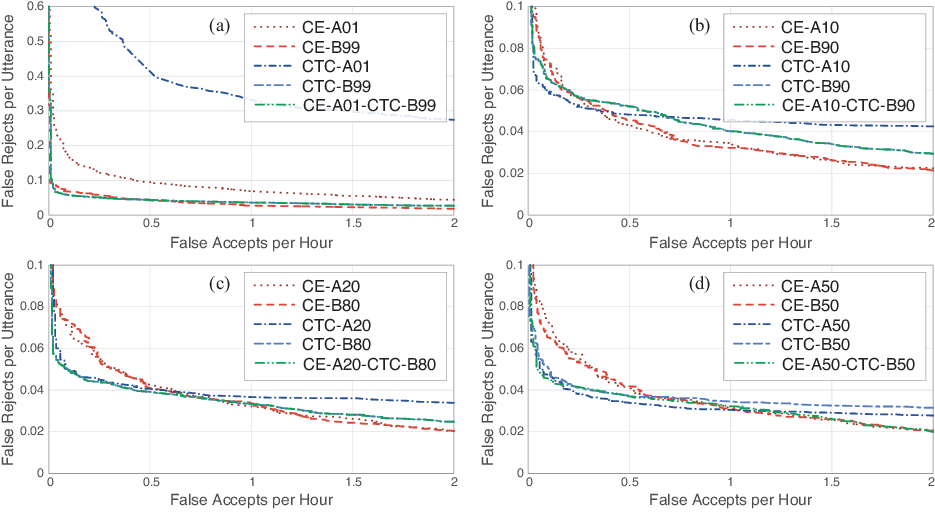}
    \caption{DET curves for the pairs \texttt{A01-B99}, \texttt{A10-B90}, \texttt{A20-B80}, and \texttt{A50-B50}, for the alignment-based, alignment-free, and hybrid alignment settings.
    }
    \label{fig:ce_ind}
\end{figure*}

\begin{figure*}[t!]
    \centering
    \includegraphics[width=\linewidth]{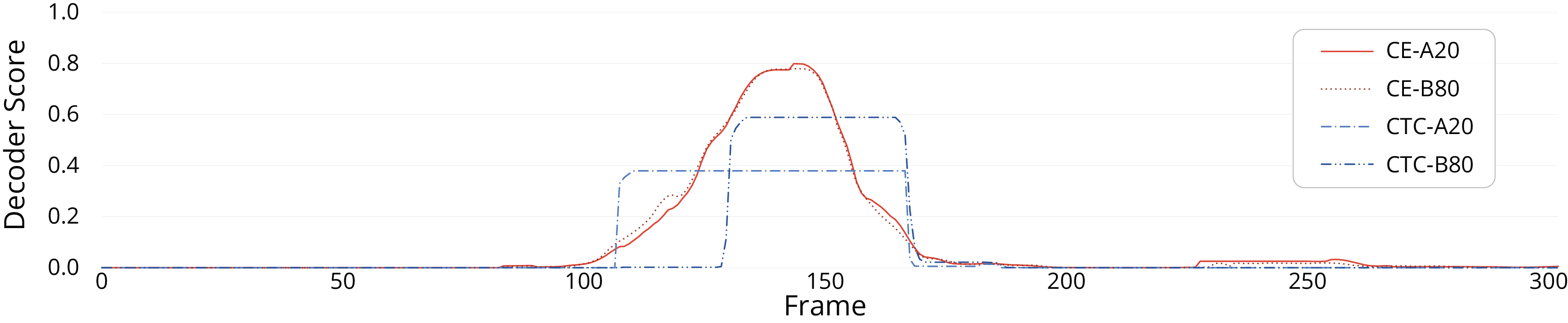}
    \caption{Decoder scores of one positive utterance for the \texttt{CE-A20}, \texttt{CE-B80}, \texttt{CTC-A20}, and \texttt{CTC-B80} models.}
    \label{fig:decoder_scores}
\end{figure*}

\autoref{fig:ce_ind} presents the DET curves for the alignment-based, alignment-free, and hybrid alignment models for each A/B split separately. \autoref{tab:FRR_01FAh} presents the FRR for each model at the $0.1$ FAh level. Note that the hybrid-alignment system only contains results for half of the settings. In these cases, the result for set \texttt{B[100-X]} refers to the hybrid model trained first with \texttt{A[X]} with cross-entropy and then with \texttt{B[100-X]} with CTC. \autoref{fig:decoder_scores} presents the decoder scores for four of our models for the same positive utterance (\texttt{CE-A20}, \texttt{CE-B80}, \texttt{CTC-A20}, and \texttt{CTC-B80}). We can observe the model triggering in the presence of the wake word. \autoref{tab:latency} presents the measured latency ($\mu \pm \sigma$) of the CTC model with respect to the equivalent cross entropy one in milliseconds.

\begin{table}
    \centering
    \caption{False Rejection Rate at the $0.1$ FAh level. For the Hybrid Alignment, the B-set trained models were initialized with the weights of its complementary A-set model.}
    \begin{tabular}{cccc}
    \toprule
    Dataset & Align.-Based & Align.-Free & Hybrid-Align.\textsuperscript{*}\\
    \midrule
    \texttt{A01}     & $16.50\%$ & $77.99\%$ & $-$ \\ 
    \texttt{A10}     &  $7.19\%$ &  $5.98\%$ & $-$ \\ 
    \texttt{A20}     &  $7.07\%$ &  $4.95\%$ & $-$ \\ 
    \texttt{A50}     &  $7.24\%$ &  $4.74\%$ & $-$ \\ 
    \texttt{B50}     &  $6.51\%$ &  $5.03\%$ & $5.69\%$ \\ 
    \texttt{B80}     &  $7.28\%$ &  $4.97\%$ & $4.97\%$ \\ 
    \texttt{B90}     &  $7.43\%$ &  $6.53\%$ & $6.53\%$ \\ 
    \texttt{B99}     &  $6.77\%$ &  $5.69\%$ & $5.69\%$ \\ 
    \bottomrule
    \end{tabular}
    \label{tab:FRR_01FAh}
\end{table}

\begin{table}
    \centering
    \caption{Latency between the cross entropy and the CTC trained models with a threshold of $40\%$ of the peak value (minimum of $0.20$). Negative values indicate that the CTC model triggered first, while positive values indicate the opposite.}
    \begin{tabular}{cccc}
    \toprule
    & & \multicolumn{2}{c}{False Rejection Rate} \\
    Dataset & Latency (ms) & CE & CTC \\
    \midrule
    \texttt{A01} & $-99.2 \pm 487.6$ & $2.7\%$ & $0.0\%$ \\
    \texttt{A10} & $-62.2 \pm 162.8$ & $1.1\%$ & $0.0\%$ \\
    \texttt{A20} & $-54.9 \pm 161.9$ & $1.0\%$ & $0.0\%$ \\
    \texttt{A50} & $-47.1 \pm 177.1$ & $1.0\%$ & $0.0\%$ \\
    \texttt{B50} & $-64.8 \pm 165.6$ & $1.0\%$ & $0.0\%$ \\
    \texttt{B80} &  $87.1 \pm  90.1$ & $1.0\%$ & $0.0\%$ \\
    \texttt{B90} & $101.5 \pm 114.9$ & $1.0\%$ & $0.0\%$ \\
    \texttt{B99} & $100.7 \pm 129.9$ & $0.9\%$ & $0.0\%$ \\
    \bottomrule
    \end{tabular}
    \label{tab:latency}
\end{table}

\section{Discussion}\label{sec:discussion}

Even though we did not confirm our initial hypothesis, this work presents many exciting findings. First, we were able to train, evaluate and compare the two traditional approaches to wake word detection, plus the proposed hybrid method. Except for \texttt{CE-A01} and \texttt{CTC-A01}, the models reached an FRR below $8\%$ at the $0.1$ FAh level, which is very encouraging given that many of the training sets are considered small.

According to our experimentation, the best alignment-based performances occurred with \texttt{B50} and \texttt{B99} -- the latter is the most extensive set -- but with only $10\%$ of the positive data (\texttt{A10}), around 23\,000 positive utterances, we were capable of training a wake word system that closely matches the user requirements and has an indistinguishable performance to its complementary set (\texttt{B90}). As \autoref{fig:ce_ind} shows, \texttt{A01} yields inferior results since most of the data observed during training are negative samples, and the model is not exposed enough to the wake word. Nonetheless, when the model is trained with only $10\%$ of the data, the alignment-based model reaches an FRR of $7.19\%$ in the $0.1$ FAh level, which is competitive with the models trained with more data, even though it is above the target ($5\%$). This is an important finding compared to previous works, which report results with a total of 1 million training utterances~\cite{alvarez2019end}. Collecting data for wake word detection incurs a tremendous cost for research institutions and organizations, especially in the initial phases of product development. Reducing the data needed for wake-word detection model training not only improves the expenses of data collection but also facilitates the documentation of the dataset and reduces GPU hours spent on training, consequently improving the carbon footprint of the systems.

Our alignment-free experiments show that the models trained with CTC improve the performance for lower levels of FAh when compared to cross-entropy. However, for FAh above $0.5$, the alignment-based model achieves a lower FRR. We observed that with only $20\%$ (about 46\,000 samples) of the data, we achieve a performance that is below the target operating point, reaching an FRR of $4.95\%$ at the $0.1$ FAh level, which is indistinguishable from the performance observed with the complementary set (\texttt{B80}). The best performance is achieved with $50\%$ (about 115\,000 samples) of the data (FRR $4.74\%$ at $0.1$ FAh). The comparison between the alignment-based and alignment-free configurations shows that it is unreasonable to claim that one approach is strictly better than the other. The choice should consider the user needs, the operating point, and the availability of resources for developing the wake word system.

The result is encouraging since unaligned data is easier to obtain. Datasets for wake word detection are available such as the SNIPS dataset for keyword spotting~\cite{coucke2019efficient}, which contains 5\,876 positive (1\,179 speakers) and 45\,344 negative (3\,330 speakers) utterances in the train set, the Mobvoi single wake word (private)~\cite{wang2019adversarial}, with 19\,684 positive and 54\,450 negative utterances for training, and the Mobvoi (SLR87)~\cite{hou2019region} datasets, with 43\,625 positive and 130\,967 negative utterances. Results reported in these sets \cite{coucke2019efficient, wang2020wake} are better than the ones presented in this study, but it is important to highlight that we use a completely different dataset for evaluation, with a different wake word, and it is not trivial to estimate how these works would perform with our data. In addition, the main outcome of our study is not beating state-of-the-art wake word detection but systematically comparing different alignment approaches using the same setup.

A secondary observation is related to the latency of CTC-based systems that have been reported in the literature~\cite{senior2015acoustic}. \autoref{fig:decoder_scores} shows an interesting behavior of the decoder scores for the two approaches. The scores for the CTC systems behave as a step function, reaching their peak as soon as the system activates. Contrarily, the cross entropy scores gradually grow until they reach their peak. In any case, it is unclear which system produces a more significant latency. \autoref{tab:latency} shows that for most of the splits, the alignment-free system triggers faster (negative values), which is probably explained by the abrupt transition between inactive and active states seen in \autoref{fig:decoder_scores}. However, the substantial standard deviations indicate that we cannot take that as a rule. Using the alignment-free approach \textit{in lieu} of the cross entropy requires further exploration for use cases where latency is a strong constraint.

To evaluate the hybrid model, we must carefully analyze \autoref{fig:ce_ind}. We observe that with the \texttt{1/99}, \texttt{10/90}, and \texttt{20/80} ratios, the hybrid training converges to the alignment-free system. Considering the relationship cross entropy and CTC have with the blank token, we can reason on top of this behavior. For cross-entropy training, the blank token does not exist; hence, the model should always emit zero probability for it. However, the blank token is critical for training CTC-based systems. The model first learns to emit blank tokens and then fills the gaps with actual ones so that the final emissions matrix will decode to the target sequence.

Nevertheless, the model benefits from both approaches when we have an aligned-to-unaligned ratio of \texttt{50/50}. It behaves like the alignment-free model for lower FAh levels, but it deviates towards the performance of the alignment-based model for higher FAh. This peculiar behavior indicates that when the model is trained with a balanced amount of data in the two phases, it can improve the performance of the two traditional approaches. However, future research and experiments are needed to understand if this result holds for different amounts of data, for example, varying the split size while holding the ratio constant.

\section{Conclusions}\label{sec:conclusions}

Most publicly available ASR datasets \cite{panayotov2015librispeech, ljspeech17, ardila2019common} do not contain alignment information. Even private datasets from industry will most likely lack phonetic alignment~\cite{wang2019adversarial, hou2019region}. For many languages other than English, a forced aligner might not exist to pre-process the data. All of these raise the relevance of alignment-free systems. Nonetheless, fine-grained phonetic annotations are still helpful for many speech applications~\cite{ribeiro2022automatic}, and the benefit from the data available is desirable. For future work, more experiments in the hybrid approach should be conducted. Moreover, a fruitful research line is to explore how to conciliate the blank token on the hybrid training. Alternatively, semi-supervised training using pseudo-labels~\cite{wang2022semi} shows up as an exciting alternative since we showed how to achieve an outstanding model with a limited amount of positive data. 

This work leaves a few learned lessons related to the wake word detection task. We tested our initial hypothesis, which was proved to be wrong, and confidently addressed the research questions proposed. Our experiments provide a reasonable estimation of the data collection needs for training wake word detection models, which is especially useful for teams that do not yet have a final product deployed in the market. In addition, to the best of our knowledge, this is the first study to evaluate and compare alignment-based and alignment-free methods for wake word detection under the same settings and resources and the first proposal of a hybrid alignment system that is compared to the traditional ones. We hope that the results of our study will benefit academia and the industry in developing more efficient voice applications in the future.

\clearpage

\bibliographystyle{IEEEtran}
\bibliography{mybib}

\begin{thebibliography}{10}
\providecommand{\url}[1]{#1}
\csname url@samestyle\endcsname
\providecommand{\newblock}{\relax}
\providecommand{\bibinfo}[2]{#2}
\providecommand{\BIBentrySTDinterwordspacing}{\spaceskip=0pt\relax}
\providecommand{\BIBentryALTinterwordstretchfactor}{4}
\providecommand{\BIBentryALTinterwordspacing}{\spaceskip=\fontdimen2\font plus
\BIBentryALTinterwordstretchfactor\fontdimen3\font minus
  \fontdimen4\font\relax}
\providecommand{\BIBforeignlanguage}[2]{{%
\expandafter\ifx\csname l@#1\endcsname\relax
\typeout{** WARNING: IEEEtran.bst: No hyphenation pattern has been}%
\typeout{** loaded for the language `#1'. Using the pattern for}%
\typeout{** the default language instead.}%
\else
\language=\csname l@#1\endcsname
\fi
#2}}
\providecommand{\BIBdecl}{\relax}
\BIBdecl

\bibitem{combs2022you}
M.~Combs, C.~Hazelwood, and R.~Joyce, ``Are you listening?--an observational
  wake word privacy study,'' \emph{Organizational Cybersecurity Journal:
  Practice, Process and People}, no. ahead-of-print, 2022.

\bibitem{schonherr2022exploring}
L.~Sch{\"o}nherr, M.~Golla, T.~Eisenhofer, J.~Wiele, D.~Kolossa, and T.~Holz,
  ``Exploring accidental triggers of smart speakers,'' \emph{Computer Speech \&
  Language}, vol.~73, p. 101328, 2022.

\bibitem{rose1990hidden}
R.~C. Rose and D.~B. Paul, ``A hidden markov model based keyword recognition
  system,'' in \emph{International Conference on Acoustics, Speech, and Signal
  Processing}.\hskip 1em plus 0.5em minus 0.4em\relax IEEE, 1990, pp. 129--132.

\bibitem{wilpon1990automatic}
J.~G. Wilpon, L.~R. Rabiner, C.-H. Lee, and E.~Goldman, ``Automatic recognition
  of keywords in unconstrained speech using hidden markov models,'' \emph{IEEE
  Transactions on Acoustics, Speech, and Signal Processing}, vol.~38, no.~11,
  pp. 1870--1878, 1990.

\bibitem{chen2014small}
G.~Chen, C.~Parada, and G.~Heigold, ``Small-footprint keyword spotting using
  deep neural networks,'' in \emph{2014 IEEE International Conference on
  Acoustics, Speech and Signal Processing (ICASSP)}.\hskip 1em plus 0.5em minus
  0.4em\relax IEEE, 2014, pp. 4087--4091.

\bibitem{yang2022lico}
H.~Yang, Z.~Yang, L.~Wan, B.~Zhang, Y.~Shi, Y.~Huang, I.~Enchev, L.~Tang,
  R.~Alvarez, M.~Sun \emph{et~al.}, ``Lico-net: Linearized convolution network
  for hardware-efficient keyword spotting,'' \emph{arXiv preprint
  arXiv:2211.04635}, 2022.

\bibitem{panchapagesan16_interspeech}
S.~Panchapagesan, M.~Sun, A.~Khare, S.~Matsoukas, A.~Mandal, B.~Hoffmeister,
  and S.~Vitaladevuni, ``{Multi-Task Learning and Weighted Cross-Entropy for
  DNN-Based Keyword Spotting},'' in \emph{Proc. Interspeech 2016}, 2016, pp.
  760--764.

\bibitem{lopez2021deep}
I.~L{\'o}pez-Espejo, Z.-H. Tan, J.~Hansen, and J.~Jensen, ``Deep spoken keyword
  spotting: An overview,'' \emph{IEEE Access}, 2021.

\bibitem{mathad2021impact}
V.~C. Mathad, T.~J. Mahr, N.~Scherer, K.~Chapman, K.~C. Hustad, J.~Liss, and
  V.~Berisha, ``The impact of forced-alignment errors on automatic
  pronunciation evaluation.'' in \emph{Interspeech}, 2021, pp. 1922--1926.

\bibitem{coucke2019efficient}
A.~Coucke, M.~Chlieh, T.~Gisselbrecht, D.~Leroy, M.~Poumeyrol, and T.~Lavril,
  ``Efficient keyword spotting using dilated convolutions and gating,'' in
  \emph{ICASSP 2019-2019 IEEE International Conference on Acoustics, Speech and
  Signal Processing (ICASSP)}.\hskip 1em plus 0.5em minus 0.4em\relax IEEE,
  2019, pp. 6351--6355.

\bibitem{hou2019region}
J.~Hou, Y.~Shi, M.~Ostendorf, M.-Y. Hwang, and L.~Xie, ``Region proposal
  network based small-footprint keyword spotting,'' \emph{IEEE Signal
  Processing Letters}, vol.~26, no.~10, pp. 1471--1475, 2019.

\bibitem{wang2019overview}
D.~Wang, X.~Wang, and S.~Lv, ``An overview of end-to-end automatic speech
  recognition,'' \emph{Symmetry}, vol.~11, no.~8, p. 1018, 2019.

\bibitem{miao2020online}
H.~Miao, G.~Cheng, P.~Zhang, and Y.~Yan, ``Online hybrid {CTC}/attention
  end-to-end automatic speech recognition architecture,'' \emph{IEEE/ACM
  Transactions on Audio, Speech, and Language Processing}, vol.~28, pp.
  1452--1465, 2020.

\bibitem{graves2012connectionist}
A.~Graves, ``Connectionist temporal classification,'' in \emph{Supervised
  sequence labelling with recurrent neural networks}.\hskip 1em plus 0.5em
  minus 0.4em\relax Springer, 2012, pp. 61--93.

\bibitem{wang2019adaptive}
S.~Wang, T.~Tuor, T.~Salonidis, K.~K. Leung, C.~Makaya, T.~He, and K.~Chan,
  ``Adaptive federated learning in resource constrained edge computing
  systems,'' \emph{IEEE Journal on Selected Areas in Communications}, vol.~37,
  no.~6, pp. 1205--1221, 2019.

\bibitem{nakkiran2015compressing}
P.~Nakkiran, R.~Alvarez, R.~Prabhavalkar, and C.~Parada, ``Compressing deep
  neural networks using a rank-constrained topology,'' 2015.

\bibitem{alvarez2019end}
R.~Alvarez and H.-J. Park, ``End-to-end streaming keyword spotting,'' in
  \emph{ICASSP 2019-2019 IEEE International Conference on Acoustics, Speech and
  Signal Processing (ICASSP)}.\hskip 1em plus 0.5em minus 0.4em\relax IEEE,
  2019, pp. 6336--6340.

\bibitem{kingma2014adam}
D.~P. Kingma and J.~Ba, ``Adam: A method for stochastic optimization,''
  \emph{arXiv preprint arXiv:1412.6980}, 2014.

\bibitem{NEURIPS2019_9015}
\BIBentryALTinterwordspacing
A.~Paszke, S.~Gross, F.~Massa, A.~Lerer, J.~Bradbury, G.~Chanan, T.~Killeen,
  Z.~Lin, N.~Gimelshein, L.~Antiga, A.~Desmaison, A.~Kopf, E.~Yang, Z.~DeVito,
  M.~Raison, A.~Tejani, S.~Chilamkurthy, B.~Steiner, L.~Fang, J.~Bai, and
  S.~Chintala, ``{PyTorch}: An imperative style, high-performance deep learning
  library,'' in \emph{Advances in Neural Information Processing Systems
  32}.\hskip 1em plus 0.5em minus 0.4em\relax Curran Associates, Inc., 2019,
  pp. 8024--8035. [Online]. Available:
  \url{http://papers.neurips.cc/paper/9015-pytorch-an-imperative-style-high-performance-deep-learning-library.pdf}
\BIBentrySTDinterwordspacing

\bibitem{martin1997det}
A.~Martin, G.~Doddington, T.~Kamm, M.~Ordowski, and M.~Przybocki, ``The {DET}
  curve in assessment of detection task performance,'' National Inst of
  Standards and Technology Gaithersburg MD, Tech. Rep., 1997.

\bibitem{wang2019adversarial}
X.~Wang, S.~Sun, C.~Shan, J.~Hou, L.~Xie, S.~Li, and X.~Lei, ``Adversarial
  examples for improving end-to-end attention-based small-footprint keyword
  spotting,'' in \emph{ICASSP 2019-2019 IEEE International Conference on
  Acoustics, Speech and Signal Processing (ICASSP)}.\hskip 1em plus 0.5em minus
  0.4em\relax IEEE, 2019, pp. 6366--6370.

\bibitem{wang2020wake}
Y.~Wang, H.~Lv, D.~Povey, L.~Xie, and S.~Khudanpur, ``Wake word detection with
  alignment-free lattice-free {MMI},'' \emph{arXiv preprint arXiv:2005.08347},
  2020.

\bibitem{senior2015acoustic}
A.~Senior, H.~Sak, F.~de~Chaumont~Quitry, T.~Sainath, and K.~Rao, ``Acoustic
  modelling with {CD-CTC-SMBR} {LSTM} {RNNS},'' in \emph{2015 IEEE Workshop on
  Automatic Speech Recognition and Understanding (ASRU)}.\hskip 1em plus 0.5em
  minus 0.4em\relax IEEE, 2015, pp. 604--609.

\bibitem{panayotov2015librispeech}
V.~Panayotov, G.~Chen, D.~Povey, and S.~Khudanpur, ``Librispeech: an {ASR}
  corpus based on public domain audio books,'' in \emph{2015 IEEE international
  conference on acoustics, speech and signal processing (ICASSP)}.\hskip 1em
  plus 0.5em minus 0.4em\relax IEEE, 2015, pp. 5206--5210.

\bibitem{ljspeech17}
K.~Ito and L.~Johnson, ``The {LJ} speech dataset,''
  \url{https://keithito.com/LJ-Speech-Dataset/}, 2017.

\bibitem{ardila2019common}
R.~Ardila, M.~Branson, K.~Davis, M.~Henretty, M.~Kohler, J.~Meyer, R.~Morais,
  L.~Saunders, F.~M. Tyers, and G.~Weber, ``Common voice: A
  massively-multilingual speech corpus,'' \emph{arXiv preprint
  arXiv:1912.06670}, 2019.

\bibitem{ribeiro2022automatic}
V.~Ribeiro, K.~Isaieva, J.~Leclere, P.-A. Vuissoz, and Y.~Laprie, ``Automatic
  generation of the complete vocal tract shape from the sequence of phonemes to
  be articulated,'' \emph{Speech Communication}, vol. 141, pp. 1--13, 2022.

\bibitem{wang2022semi}
Y.~Wang, H.~Wang, Y.~Shen, J.~Fei, W.~Li, G.~Jin, L.~Wu, R.~Zhao, and X.~Le,
  ``Semi-supervised semantic segmentation using unreliable pseudo-labels,'' in
  \emph{Proceedings of the IEEE/CVF Conference on Computer Vision and Pattern
  Recognition}, 2022, pp. 4248--4257.

\end{thebibliography}

\end{document}